\documentclass[10pt,twocolumn]{article}

\usepackage[T1]{fontenc}
\usepackage{lmodern} 

\usepackage{amsmath,amssymb,amsthm}

\usepackage{graphicx}
\usepackage{float}
\usepackage{booktabs}
\usepackage{multirow}
\usepackage{tabularx}
\usepackage{array}
\newcolumntype{Y}{>{\raggedright\arraybackslash}X}
\newcolumntype{Z}{>{\centering\arraybackslash}X}
\newcolumntype{W}{>{\raggedleft\arraybackslash}X}
\renewcommand{\arraystretch}{1.3}

\usepackage{subcaption}
\usepackage{algorithm}
\usepackage{algpseudocode}

\usepackage{enumitem}
\setlist{topsep=3pt, itemsep=3pt, parsep=0pt, partopsep=0pt}

\setlist[enumerate,1]{label=\alph*), leftmargin=*}

\usepackage[numbers,sort&compress]{natbib}

\usepackage[colorlinks=true,
linkcolor=blue,
citecolor=blue,
urlcolor=blue]{hyperref}

\usepackage{geometry}
\usepackage{titlesec}
\titleformat{\section}
{\normalsize\bfseries}
{\thesection.}
{0.5em}
{}
\titlespacing*{\section}{0pt}{2.5ex plus 0.2ex minus 0.2ex}{1.0ex}

\titleformat{\subsection}
{\normalsize\itshape}
{\thesubsection}
{0.5em}
{}
\titlespacing*{\subsection}{0pt}{1.5ex plus 0.2ex minus 0.2ex}{0.8ex}

\titleformat{\subsubsection}
{\normalsize\itshape}
{\thesubsubsection}
{0.6em}
{}
\titlespacing*{\subsubsection}{0pt}{1.2ex plus 0.2ex minus 0.2ex}{0.6ex}

\titleformat{\paragraph}
{\normalsize\bfseries}
{}
{0em}
{}

\titlespacing*{\paragraph}
{0pt}        
{1.2ex}      
{0.2em}      

\usepackage[font=small,labelfont=bf]{caption}
\DeclareMathOperator{\I}{I}

\usepackage[dvipsnames]{xcolor}
\usepackage[normalem]{ulem}
\usepackage{soul}
\sethlcolor{yellow!35}

\soulregister\cite7
\soulregister\citet7
\soulregister\citep7
\soulregister\ref7
\soulregister\pageref7

\usepackage{framed}
\definecolor{shadecolor}{rgb}{1,1,0.65}
\definecolor{mygrey}{gray}{0.965}

\title{Bayesian Deployment Approval for Learned Landing Controllers under Finite Rollout Validation}

\author{
	Fei Jiang\thanks{
		Independent Researcher, Seattle, WA, USA. 
		Correspondence to: jiangfeicq@gmail.com.
		Fei Jiang and Lei Yang contributed equally to this work.
	}
	\and
	Lei Yang\footnotemark[1]
}

\date{}
\begin{document}
\maketitle
\begin{abstract}
	Reinforcement learning and data-driven autonomous controllers are commonly evaluated using cumulative reward and empirical success frequency under finite simulation trajectories. However, such empirical metrics do not necessarily provide sufficient statistical evidence regarding deployment readiness under uncertainty. This work develops a Bayesian approval framework for learned autonomous landing controllers under finite rollout evidence. A probabilistic landing capability formulation is introduced based on touchdown safety satisfaction under uncertain operating conditions, while Bayesian posterior inference is used to quantify uncertainty regarding the true deployment capability of learned policies. Posterior approval probability and posterior deployment risk are further introduced for deployment-oriented evaluation, together with a sequential validation framework supporting approve/reject/continue decisions during progressive rollout testing. Simulation experiments using PPO and SAC controllers demonstrate that empirical success and reward optimization may produce overconfident deployment interpretation under limited validation evidence, whereas posterior approval inference provides a more uncertainty-calibrated assessment of deployment readiness. The proposed framework provides a practical statistical connection between conventional reinforcement-learning evaluation and deployment-oriented validation under uncertainty and may be generalized to broader classes of learned autonomous systems.
\end{abstract}


\setcounter{tocdepth}{2}  
\section{Introduction}
\label{sec:introduction}

Reinforcement learning (RL) and data-driven control methods have demonstrated strong capability in autonomous decision-making tasks involving nonlinear dynamics, uncertainty, and sequential control \cite{sutton1998reinforcement,mnih2015human}. 
Recent advances in simulation-based learning have enabled learned controllers to achieve high empirical performance in robotics, autonomous driving, unmanned aerial vehicles, and autonomous landing systems \cite{schulman2017proximal,haarnoja2018soft}. 
In particular, autonomous landing remains a challenging control problem due to coupled nonlinear dynamics, environmental disturbances, state uncertainty, and strict touchdown safety requirements. 
Modern learning-based controllers are typically trained using large numbers of simulation trajectories and subsequently evaluated using empirical metrics such as cumulative reward, average return, or landing success frequency.

Despite these advances, an important gap remains between empirical controller performance and deployment-oriented validation under uncertainty \cite{amodei2016concrete}. Most existing RL-based landing studies primarily report reward progression or observed success frequency over a finite number of rollout episodes \cite{koch2019reinforcement,rodriguez2019deep}. However, empirical performance alone does not necessarily provide sufficient statistical evidence regarding whether a learned controller is suitable for deployment under uncertain operating conditions. A controller achieving high observed success during validation may still exhibit substantial uncertainty regarding its true operational capability, particularly when validation evidence is limited or when rare unsafe events remain insufficiently sampled. This problem is closely related to broader challenges in safe reinforcement learning, uncertainty calibration, and AI deployment safety \cite{garcia2015comprehensive,koopman2017autonomous,kahn2017uncertainty}. Furthermore, reward optimization and deployment readiness represent different evaluation objectives. A policy with larger cumulative reward may still produce unsafe touchdown behavior under certain disturbances or unseen operating conditions \cite{alshiekh2018safe}. Consequently, deployment decisions based solely on empirical reward or observed success frequency may lead to overconfident approval of insufficiently validated learned controllers.

The distinction between empirical performance and statistically grounded deployment confidence has also been emphasized in recent literature on uncertainty-aware learning-based systems. Learned models and controllers may exhibit overconfident behavior under finite observations or distribution shift conditions \cite{guo2017calibration,lakshminarayanan2017simple,ovadia2019can}. Although uncertainty-aware reinforcement learning and safe-control formulations have received increasing attention \cite{kahn2017uncertainty, garcia2015comprehensive}, comparatively limited work has investigated finite-sample deployment approval and statistical validation for learned autonomous controllers operating under uncertain environments. Recent studies have additionally explored probabilistically justified decision frameworks and finite-sample risk-aware approval formulations under limited observations \cite{jiang2026finite,jiang2026risk,jiang2026machine}.
However, deployment-oriented statistical approval mechanisms for learning-based autonomous landing systems remain insufficiently developed.

This problem becomes particularly important in safety-critical autonomous systems.
In practical deployment, the key question is not merely whether a controller succeeds in observed simulations, but whether the available rollout evidence is sufficient to justify deployment under uncertainty.
This leads to a finite-sample statistical decision problem for learned autonomous policies: how should deployment capability and approval confidence be quantified from limited rollout observations?
These questions are closely related to Bayesian inference, sequential hypothesis testing, and statistical decision theory \cite{wald1992sequential,gelman1995bayesian,berger2013statistical,tartakovsky2014sequential}.
Related concepts also arise in reliability engineering, conformity assessment, and uncertainty-aware industrial decision systems \cite{pendrill2014using,modarres2016reliability,jcgm106}.

This work develops a Bayesian deployment approval framework for learned landing controllers using finite rollout validation evidence. Instead of evaluating controllers solely through empirical reward or average success frequency, deployment capability is defined as the probability that a landing trajectory satisfies prescribed touchdown safety constraints under randomized operating conditions. A safe landing event is characterized using multiple engineering criteria, including touchdown position error, vertical touchdown speed, pitch stability, and related safety constraints.

Building upon this formulation, rollout trajectories are modeled as Bernoulli safety outcomes, and posterior uncertainty regarding deployment capability is quantified through Bayesian updating \cite{gelman1995bayesian}.
The framework further introduces posterior approval probability and false-approval risk for deployment-oriented controller evaluation. To improve validation efficiency, a sequential validation mechanism is developed to support approve/reject/continue decisions during progressive rollout testing \cite{wald1992sequential,tartakovsky2014sequential}. The resulting formulation enables deployment-oriented statistical approval analysis under finite rollout uncertainty.

The proposed methodology is intentionally developed independently of the underlying learning algorithm.
The controller layer may consist of reinforcement learning policies, heuristic controllers, classical feedback systems, or hybrid architectures. Accordingly, the primary objective of this work is to investigate deployment-oriented statistical validation under finite rollout uncertainty and to establish a statistically calibrated framework for evaluating deployment capability under finite validation evidence.

Simulation studies are conducted using autonomous landing environments with randomized disturbances and uncertain initial conditions.
Multiple controllers are evaluated to investigate posterior approval evolution, finite-sample uncertainty, reward--approval mismatch, and sequential validation efficiency.
The results demonstrate that empirical reward alone may provide misleading deployment interpretation, whereas the proposed Bayesian framework yields a more posterior-aware assessment of deployment readiness under finite rollout evidence.

The main contributions are summarized as follows:

\begin{enumerate}
	\item A landing capability formulation is developed for learned autonomous controllers based on the probability of satisfying multiple touchdown safety constraints under uncertainty.
	
	\item A Bayesian finite-sample deployment approval framework is proposed to estimate posterior controller capability using rollout validation trajectories.
	
	\item Posterior approval probability and posterior false-approval risk are introduced for deployment-oriented evaluation of learned landing policies.
	
	\item A sequential validation mechanism is developed to support approve/reject/continue decisions during progressive rollout testing.
	
	\item Simulation studies demonstrate the distinction between reward optimization, empirical landing success, and uncertainty-calibrated deployment approval under finite validation evidence.
\end{enumerate}

The remainder of this paper is organized as follows.
Section~\ref{sec:problem} introduces the autonomous landing environment, touchdown safety constraints, and rollout validation formulation.
Section~\ref{sec:capability} develops the probabilistic landing capability formulation under uncertain operating conditions.
Section~\ref{sec:bayesian} presents the Bayesian finite-sample deployment approval framework, including posterior approval probability and false-approval risk.
Section~\ref{sec:sequential} develops the sequential approve/reject/continue validation framework under progressive rollout evidence.
Section~\ref{sec:simulation} describes the simulation environment, controller training procedures, and deployment-validation protocol.
Section~\ref{sec:results} presents experimental results on posterior approval evolution, reward--approval mismatch, and finite-sample deployment uncertainty.
Section~\ref{sec:discussion} discusses deployment-oriented interpretation, operating-distribution dependence, and future extensions.
Finally, concluding remarks are provided in Section~\ref{sec:conclusion}.

 \section{Problem Formulation}
 \label{sec:problem}
 \begin{figure*}[!t]
 	\centering
 	\includegraphics[width=1.00\textwidth]
 	{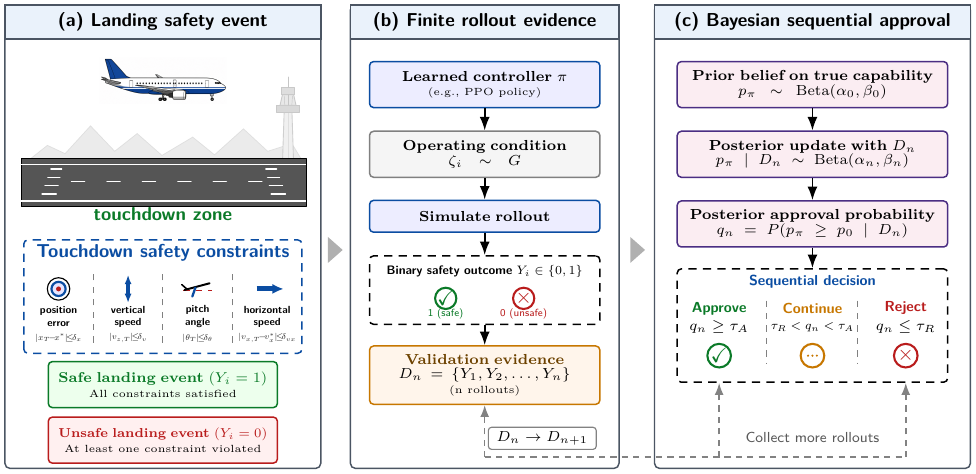}
 	\caption{
 		Overview of the proposed Bayesian deployment approval framework for autonomous landing validation under finite rollout evidence. 		
 		(a) Define a safe touchdown event based on multiple touchdown safety constraints.
 		(b) Under sampled operating conditions, each rollout generates a binary outcome $Y_i$ and contributes to the evidence set $D_n$. 		
 		(c) Bayesian inference quantifies deployment capability $p_\pi$; deployment decisions use the posterior approval probability $q_n$ with thresholds $\tau_A$ and $\tau_R$. 		
 		Touchdown variables include position \(x_T\), vertical speed \(v_{z,T}\), pitch angle \(\theta_T\), horizontal speed \(v_{x,T}\), and touchdown-contact indicators.
 	}
 	\label{fig:framework_overview}
 \end{figure*}
 
 Figure~\ref{fig:framework_overview} provides an overview of the proposed Bayesian deployment approval framework for learned autonomous landing controllers under uncertainty. 
 A rollout trajectory is first evaluated using multiple touchdown safety constraints to determine whether a safe landing event occurs. 
 Under randomized operating conditions, rollout validation generates binary safety outcomes that are subsequently aggregated into a finite evidence set. 
 Bayesian posterior updating is then used to quantify deployment approval confidence and support sequential approve/reject/continue decisions under finite rollout evidence.
 \subsection{Autonomous Landing Environment}
 \label{subsec:environment}
 
 Consider an autonomous landing environment in which a learned controller guides an aircraft-like vehicle toward a target touchdown region under uncertain operating conditions. The environment is modeled as a discrete-time dynamical system
 \[
 s_{t+1}=f(s_t,u_t,\xi_t),
 \]
 
 where \(s_t\in\mathcal{S}\) denotes the system state at time step \(t\), \(u_t\in\mathcal{U}\) denotes the control action, and \(\xi_t\) represents stochastic disturbances and modeling uncertainty. 
 The disturbance term may include environmental perturbations such as wind variation, sensor noise, initial-condition uncertainty, actuator variability, or other stochastic effects.
 
 For a simplified two-dimensional landing environment, the state vector is represented as
\begin{equation}
	s_t=
	(x_t,z_t,v_{x,t},v_{z,t},\theta_t,\omega_t),
	\label{eq:state_vector}
\end{equation}
  where \(x_t\) and \(z_t\) denote horizontal position and altitude, \(v_{x,t}\) and \(v_{z,t}\) denote horizontal and vertical velocities, \(\theta_t\) denotes pitch angle, and \(\omega_t\) denotes pitch angular rate. 
 The control input is represented as
 \[
 u_t=(T_t,\delta_t),
 \]
 where \(T_t\) denotes throttle-related thrust control and \(\delta_t\) denotes elevator or attitude-control input.
 
 The present work intentionally adopts a simplified landing environment in order to focus on statistical reliability approval under uncertainty rather than high-fidelity aircraft dynamics. 
 The proposed framework is independent of the specific learning architecture and may be applied to reinforcement learning controllers, heuristic policies, classical feedback controllers, or hybrid control systems.
 
 A landing trajectory generated by policy $\pi$ is denoted by
 \begin{equation}
 	\tau =
 	(s_0,u_0,s_1,u_1,\dots,s_T),
 	\label{eq:trajectory}
 \end{equation}
where \(T\) denotes the terminal touchdown time. The policy may be deterministic,
 \[
 u_t=\pi(s_t),
 \]
 or stochastic,
 \[
 u_t\sim \pi(\cdot\mid s_t).
 \]
 
 The initial state is sampled from an operating-condition distribution $ s_0\sim \rho_0$,
 which represents uncertainty in initial altitude, speed, orientation, or environmental conditions.
 
 \subsection{Landing Safety Constraints}
 \label{subsec:safety}
 
 The objective of autonomous landing is not merely to maximize cumulative reward, but to satisfy predefined touchdown safety requirements. 
 A landing trajectory is considered safe only if multiple engineering constraints are simultaneously satisfied at touchdown.
 
Let \(x^\ast\) denote the desired touchdown position. 
A safe landing event is defined as

\begin{equation}
	\mathcal{S}
	=
	\left\{
	\begin{array}{l}
		|x_T-x^\ast|\le \delta_x\\
		|v_{z,T}|\le \delta_v\\
		|\theta_T|\le \delta_\theta\\
		|v_{x,T}-v_x^\ast|\le \delta_{vx}\\
		c_T^{\mathrm{left}}\ge \delta_c\\
		c_T^{\mathrm{right}}\ge \delta_c
	\end{array}
	\right.,
	\label{eq:safe_landing_event}
\end{equation}

where 
\begin{itemize}
	\item \(\delta_x\) denotes the allowable touchdown position error,
	\item \(\delta_v\) denotes the allowable vertical touchdown speed,
	\item \(\delta_\theta\) denotes the allowable pitch-angle deviation,
	\item \(\delta_{vx}\) denotes the allowable horizontal touchdown speed,
	\item \(c_T^{\mathrm{left}}\) and \(c_T^{\mathrm{right}}\) denote the normalized touchdown-contact indicators for the left and right landing legs,
	\item \(\delta_c\) denotes the minimum touchdown-contact threshold.
\end{itemize}

The safety event formulation emphasizes that successful landing is inherently a multi-constraint reliability problem. 
A trajectory may achieve high cumulative reward while still violating one or more touchdown safety constraints. 
Consequently, empirical reward optimization alone does not necessarily imply reliable landing safety under uncertainty.
 
For convenience, the safe landing event may also be represented as the intersection of multiple constraint events,
\begin{equation}
	\mathcal{S}
	=
	\bigcap_{j=1}^{m}\mathcal{S}_j,
	\label{eq:joint_safe_event}
\end{equation}
 where each \(\mathcal{S}_j\) corresponds to an individual touchdown safety requirement.
 
\subsection{Validation Trajectories and Safety Outcomes}
\label{subsec:validation}

After policy training is completed, the learned controller is evaluated using a set of rollout validation episodes generated under randomized operating conditions. 
Each rollout trajectory \(\tau_i\) produces a corresponding binary safety outcome
\begin{equation}
	Y_i=
	\begin{cases}
		1, & \tau_i\in\mathcal{S},\\
		0, & \tau_i\notin\mathcal{S},
	\end{cases}
	\qquad i=1,\dots,n,
	\label{eq:bernoulli_safety_event}
\end{equation}
where \(Y_i=1\) indicates that all touchdown safety constraints are satisfied, while \(Y_i=0\) indicates that at least one safety constraint is violated.

Under the assumed operating-condition distribution, the rollout validation outcomes are modeled as conditionally independent Bernoulli random variables given the unknown deployment capability \(p_\pi\). Under independently sampled rollout initialization and disturbance conditions,
\begin{equation}
	Y_i\mid p_\pi
	\overset{\mathrm{i.i.d.}}{\sim}
	\mathrm{Bernoulli}(p_\pi),
	\label{eq:bernoulli_model}
\end{equation}
where \(p_\pi\) denotes the probability that policy \(\pi\) achieves a safe landing under uncertain operating conditions.

This assumption is appropriate when rollout validation episodes are generated using independently sampled initial conditions, disturbances, and random seeds; correlated or adaptive rollout-generation mechanisms would require extended dependence-aware likelihood formulations.

The validation dataset is represented by
\begin{equation}
	D_n=\{Y_1,Y_2,\dots,Y_n\},
	\label{eq:evidence_set}
\end{equation}
with
\begin{equation}
	S_n=\sum_{i=1}^{n}Y_i,
	\qquad
	F_n=n-S_n,
	\qquad
	\hat p_n=\frac{S_n}{n},
	\label{eq:empirical_success_rate}
\end{equation}
denoting the numbers of successful and unsuccessful rollout trajectories together with the empirical landing success rate, respectively.

The statistical objective of the proposed framework is to determine whether the unknown deployment capability \(p_\pi\) exceeds a prescribed deployment-reliability requirement using finite rollout validation evidence. 
However, the empirical success rate alone does not fully characterize uncertainty regarding the true deployment capability under finite validation samples. 
For example, observing \(10/10\) successful landings and observing \(200/200\) successful landings both yield empirical success rate equal to one, despite substantially different levels of statistical confidence regarding deployment readiness. 
Consequently, uncertainty-aware posterior inference is required for finite-sample deployment assessment.

\section{Landing Capability under Uncertainty}
\label{sec:capability}

\subsection{Landing Capability of Learned Policies}
\label{subsec:capability_definition}

The validation outcomes defined in Section~\ref{sec:problem} characterize whether a landing trajectory satisfies the required touchdown safety constraints. 
Building upon this formulation, the reliability of a learned landing controller is quantified through a probabilistic landing capability measure.

For a policy $\pi$, the landing capability is defined as the probability that the generated landing trajectory satisfies the safe landing event,
\[
p_\pi
=
P(\tau\in\mathcal{S}\mid \pi),
\]
where \(\mathcal{S}\) denotes the safe landing event defined in \eqref{eq:joint_safe_event}. 

The probability is taken with respect to the operating-condition distribution, environmental disturbances, stochastic dynamics, and possible policy randomness.

More explicitly, the landing capability is written as
\begin{equation}
	p_\pi
	=
	\int
	\I\{\tau\in\mathcal{S}\}
	\,
	p(\tau\mid \pi,\rho_0,\Xi)
	\,d\tau,
	\label{eq:landing_capability_integral}
\end{equation}
where $\I\{\cdot\}$ denotes the indicator function, $\rho_0$ denotes the initial-condition distribution, and \(\Xi\) denotes the disturbance distribution governing stochastic operating conditions.

The parameter $p_\pi$ represents the true safety reliability of the learned controller under uncertain operating conditions. Unlike empirical success rate computed from finite validation trajectories, $p_\pi$ is an unknown population-level capability parameter characterizing the inherent deployment reliability of the policy.

The proposed formulation intentionally separates controller learning from reliability evaluation. 
The learning process generates a policy $\pi$, whereas the present framework focuses on estimating whether the resulting policy possesses sufficient landing capability for reliable deployment.

\subsection{Multi-Constraint Safety Capability}
\label{subsec:multiconstraint}

Autonomous landing safety is inherently a multi-constraint reliability problem. 
A landing trajectory may satisfy certain touchdown criteria while violating others. 
For example, a trajectory may achieve accurate touchdown position while simultaneously exhibiting excessive vertical touchdown speed or unstable pitch angle. 
Consequently, reliable landing cannot be characterized adequately using a single scalar reward or isolated performance metric.

The safe landing event is represented as the intersection of multiple constraint events as defined in \eqref{eq:joint_safe_event}, where each $\mathcal{S}_j$ denotes an individual safety constraint. 
Typical constraints include touchdown position error, touchdown vertical speed, pitch-angle stability, horizontal speed regulation, glide-path deviation, or other operational requirements.

Accordingly, the landing capability becomes
\[
p_\pi
=
P\left(
\bigcap_{j=1}^{m}\mathcal{S}_j
\mid \pi
\right).
\]

The formulation emphasizes that landing reliability depends on the joint satisfaction of all safety constraints rather than isolated marginal performance measures. 
In general, the individual constraint events are statistically dependent due to the coupled nature of landing dynamics. 
Therefore,
\begin{equation*}
	P\left(
	\bigcap_{j=1}^{m}\mathcal{S}_j
	\mid \pi
	\right)
	\neq
	\prod_{j=1}^{m}
	P(\mathcal{S}_j\mid \pi),
\end{equation*}
unless independence assumptions are explicitly imposed.

For diagnostic purposes, individual marginal safety capabilities may also be defined as
\[
p_{\pi,j}
=
P(\mathcal{S}_j\mid \pi),
\qquad j=1,\dots,m.
\]

These marginal quantities characterize the probability that individual constraints are satisfied, whereas the joint landing capability \(p_\pi\) characterizes overall deployment reliability.

\subsection{Capability under Operating Uncertainty}
\label{subsec:uncertainty}

The reliability of a learned controller depends strongly on the operating conditions under which validation is performed. 
In practical deployment scenarios, autonomous landing may be affected by environmental disturbances, sensor uncertainty, model mismatch, actuator variation, and uncertainty in initial conditions. 
Therefore, landing capability must be interpreted relative to a specified operating-condition distribution.

Let $\zeta\sim G$ denote a random operating condition sampled from distribution \(G\). 
The operating condition may include wind disturbances, initial altitude variation, initial velocity perturbations, sensor noise, or other uncertain environmental factors.

Conditioned on a fixed operating condition \(\zeta\), the conditional landing capability is defined as
\[
p_\pi(\zeta)
=
P(\tau\in\mathcal{S}\mid \pi,\zeta)
\]

The overall robust landing capability under uncertain operating conditions is then defined as
\[
p_\pi^{\mathrm{rob}}
=
P(\tau\in\mathcal{S}\mid \pi,\zeta\sim G).
\]

Equivalently,
\[
p_\pi^{\mathrm{rob}}
=
E_{\zeta\sim G}
\left[
p_\pi(\zeta)
\right].
\]

This formulation highlights that controller reliability is distribution-dependent. 
A policy validated under one operating-condition distribution may exhibit substantially different reliability under another disturbance distribution. 
Consequently, deployment-oriented reliability assessment requires explicit specification of the uncertainty distribution used during validation.

The present work focuses on statistical reliability approval under a specified operating-condition distribution rather than worst-case formal verification. 
The proposed framework therefore provides probabilistic reliability assessment conditioned on the assumed validation environment and disturbance model.

\subsection{Reward Optimization versus Reliability Capability}
\label{subsec:reward_vs_capability}

Most reinforcement learning algorithms optimize a cumulative reward objective of the form
\begin{equation}
	J(\pi)
	=
	E
	\left[
	\sum_{t=0}^{T}
	\gamma^t r_t
	\right],
	\label{eq:rl_objective}
\end{equation}
where \(r_t\) denotes the instantaneous reward and \(\gamma\in(0,1)\) denotes the discount factor. 
Although reward optimization provides an effective learning mechanism, the resulting reward value does not necessarily correspond directly to deployment reliability.

The proposed landing capability framework distinguishes reward optimization from probabilistic safety capability. 
A policy with larger cumulative reward may still exhibit lower safe-landing reliability under uncertainty. 
Conversely, a policy with slightly lower average reward may produce substantially more stable and reliable touchdown behavior.

Formally, two policies \(\pi_1\) and \(\pi_2\) may satisfy
\[
J(\pi_1)>J(\pi_2),
\]
while simultaneously satisfying $p_{\pi_1}<p_{\pi_2}$.

This mismatch occurs because reward optimization primarily reflects trajectory-level utility accumulation, whereas landing capability reflects probabilistic satisfaction of deployment-oriented safety constraints.

The distinction becomes particularly important under finite validation evidence. 
Observed empirical reward or finite-sample success rate alone may produce overconfident reliability conclusions regarding learned controllers. 
The following sections therefore develop a Bayesian finite-sample deployment approval framework for estimating posterior uncertainty regarding landing capability under limited validation trajectories.

\subsection{Deployment Validation versus Conventional RL Evaluation}
\label{subsec:deployment_vs_rl}

Conventional reinforcement learning evaluation is primarily training-oriented and typically relies on cumulative reward, average return, or empirical success frequency obtained from finite rollout experiments. Although such metrics are useful for measuring learning progress, they do not necessarily provide statistically calibrated evidence regarding deployment readiness under uncertainty.

The proposed framework distinguishes empirical rollout evaluation from deployment-oriented statistical validation. Conventional RL evaluation focuses primarily on quantities such as the reward objective in \eqref{eq:rl_objective} or the empirical success frequency in \eqref{eq:empirical_success_rate}, whereas the present framework focuses on posterior deployment confidence regarding whether the unknown landing capability satisfies the required reliability threshold (Section~\ref{subsec:approval_probability}).

These quantities serve fundamentally different purposes. Reward optimization characterizes expected trajectory-level performance during learning, whereas deployment approval characterizes reliability-aware confidence regarding safe operational capability under finite rollout evidence.

Consequently, strong empirical reward or observed landing success alone should not be interpreted as equivalent to deployment readiness. The proposed framework may therefore be viewed as a deployment-oriented extension of conventional reinforcement learning evaluation through Bayesian posterior inference and sequential validation under finite-sample uncertainty.

\begin{figure*}[!t]
	\centering
	\includegraphics[width=1.00\textwidth]
	{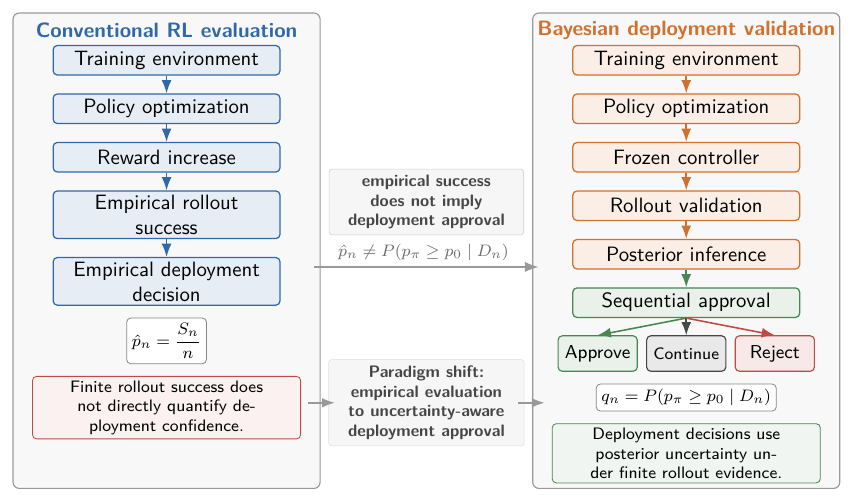}
	\caption{
		Conceptual comparison between conventional reinforcement-learning evaluation and the proposed Bayesian deployment-oriented validation framework under finite rollout uncertainty. 
		The conventional paradigm primarily emphasizes reward optimization and empirical rollout success during training, whereas the proposed framework separates controller learning from deployment approval through independent rollout validation and Bayesian posterior inference. 
		The figure further illustrates that empirical rollout success alone does not necessarily imply deployment approval under finite validation evidence, motivating uncertainty-aware deployment validation for learned autonomous systems.
	}
	\label{fig:deployment_paradigm_comparison}
\end{figure*}

Figure~\ref{fig:deployment_paradigm_comparison} illustrates the fundamental distinction between conventional reinforcement-learning evaluation and deployment-oriented statistical validation under finite rollout uncertainty. 

Conventional reinforcement learning primarily evaluates learning progress through reward optimization and empirical rollout success during training. Although such metrics are useful for policy improvement, they do not directly quantify uncertainty-calibrated evidence regarding deployment readiness under finite validation samples.

In contrast, the proposed framework separates controller learning from deployment approval by introducing an explicit deployment-validation stage based on independent rollout evidence and Bayesian posterior inference. Under this perspective, deployment approval is treated as a statistical decision problem rather than a direct consequence of empirical success frequency alone. Consequently, strong empirical performance does not necessarily imply reliable deployment capability under finite-sample uncertainty.

\section{Bayesian Deployment Approval under Finite Validation}
\label{sec:bayesian}

Given the rollout-validation formulation introduced previously, Bayesian posterior inference is now developed to quantify uncertainty regarding the unknown deployment capability of learned landing controllers under finite rollout evidence. The proposed framework further introduces posterior approval probability and posterior false-approval risk for deployment-oriented controller evaluation under uncertainty.

\subsection{Bayesian Posterior Capability Inference}
\label{subsec:posterior}

To quantify uncertainty regarding the unknown landing capability, a Bayesian inference framework is adopted. 
The landing capability \(p_\pi\) is modeled using a Beta prior distribution,
\[
p_\pi
\sim
\mathrm{Beta}(\alpha_0,\beta_0),
\]
where \(\alpha_0>0\) and \(\beta_0>0\) denote prior hyperparameters. 
The Beta prior is conjugate to the Bernoulli likelihood and therefore yields a closed-form posterior distribution.

Conditioned on the validation dataset \(D_n\) defined in \eqref{eq:evidence_set}, the likelihood function is
\[
P(D_n\mid p_\pi)
=
p_\pi^{S_n}
(1-p_\pi)^{F_n}.
\]

Combining the likelihood with the Beta prior yields the posterior distribution
\begin{equation}
	p_\pi\mid D_n
	\sim
	\mathrm{Beta}
	(
	\alpha_0+S_n,
	\beta_0+F_n
	),
	\label{eq:beta_posterior}
\end{equation}

The posterior mean landing capability is therefore
\[
E[p_\pi\mid D_n]
=
\frac{\alpha_0+S_n}
{\alpha_0+\beta_0+n}.
\]

Similarly, the posterior variance is
\[
\mathrm{Var}(p_\pi\mid D_n)
=
\frac{
	(\alpha_0+S_n)
	(\beta_0+F_n)
}{
	(\alpha_0+\beta_0+n)^2
	(\alpha_0+\beta_0+n+1)
}.
\]

In the absence of strong prior information, a weakly informative prior such as $\alpha_0=\beta_0=1$ may be adopted, corresponding to a uniform prior over the interval \([0,1]\).

The posterior formulation provides a probabilistic characterization of uncertainty regarding the true landing capability under finite validation evidence. 
As the number of validation trajectories increases, posterior uncertainty decreases and the posterior distribution progressively concentrates around the true capability value.

\subsection{Posterior Approval Probability}
\label{subsec:approval_probability}

Suppose that deployment requires the landing capability of a controller to exceed a prescribed reliability threshold $p_0\in(0,1)$. For example, the deployment reliability threshold \(p_0=0.95\) is selected as a representative high-reliability requirement for safety-critical autonomous operation, corresponding to a requirement that the controller achieve safe landing with probability at least \(95\%\) under the assumed operating-condition distribution.

Rather than relying solely on empirical success rate, the proposed framework evaluates deployment reliability using the posterior probability that the unknown landing capability exceeds the required threshold. 
The posterior approval probability is defined as
\begin{equation}
	q_n
	=
	P(p_\pi\ge p_0\mid D_n),
	\label{eq:posterior_approval_probability}
\end{equation}

Using the Beta posterior distribution, the approval probability becomes
\[
q_n
=
1-
F_{\mathrm{Beta}}
\left(
p_0;
\alpha_0+S_n,
\beta_0+F_n
\right),
\]
where \(F_{\mathrm{Beta}}(\cdot;a,b)\) denotes the cumulative distribution function of a Beta distribution with parameters \(a\) and \(b\).

The quantity \(q_n\) measures the posterior confidence that the learned controller satisfies the required deployment reliability threshold. Unlike empirical success rate, the posterior approval probability explicitly incorporates finite-sample uncertainty.

For illustration, two controllers may both achieve empirical success rate equal to one, while exhibiting substantially different approval probabilities due to different validation sample sizes. 
Consequently, posterior approval probability provides a more calibrated reliability metric for deployment-oriented controller evaluation. 

Unlike conventional lower-confidence-bound acceptance rules, the proposed posterior approval formulation directly quantifies deployment confidence as a probabilistic exceedance event under finite rollout uncertainty \cite{berger2013statistical,wald1992sequential}.

\subsection{Posterior Deployment Risk}
\label{subsec:false_approval}

A key concern in deployment-oriented controller validation is the possibility of approving an insufficiently reliable policy due to finite validation evidence. 
To quantify this uncertainty, the proposed framework introduces posterior false-approval risk.

Suppose deployment approval requires
\begin{equation}
	p_\pi\ge p_0.
	\label{eq:deployment_requirement}
\end{equation}

If a controller is approved despite possessing true landing capability below the required threshold, a false approval event occurs. 
Conditioned on the observed validation evidence, the posterior false-approval risk is defined as
\[
r_n^{\mathrm{FA}}
=
P(p_\pi<p_0\mid D_n).
\]

Using the definition of posterior approval probability, the false-approval risk becomes
\[
r_n^{\mathrm{FA}}
=
1-q_n.
\]

Therefore, large posterior approval probability corresponds directly to small posterior false-approval risk.

A deployment-oriented approval rule may now be introduced. 
Given a prescribed approval confidence threshold \(\tau_A\in(0,1)\), a controller is approved if
\[
q_n\ge \tau_A.
\]

Equivalently, the approval condition is written as
\[
r_n^{\mathrm{FA}}
\le
1-\tau_A.
\]

For example, setting $\tau_A=0.95$ requires the posterior probability of satisfying the deployment reliability threshold to be at least \(95\%\).

The proposed framework therefore transforms controller validation into a probabilistic reliability approval problem under finite rollout validation evidence. Instead of relying solely on observed empirical success, deployment decisions are based on posterior uncertainty regarding the unknown landing capability of the learned controller. The resulting approval probabilities should therefore be interpreted as uncertainty-calibrated deployment confidence under the assumed validation distribution rather than formal deterministic certification guarantees.

The present framework focuses primarily on false-approval risk because approving an insufficiently reliable controller may lead to more severe deployment consequences in safety-critical autonomous systems. Although false rejection may increase validation conservatism or deployment delay, its treatment is left for future decision-theoretic extensions.

\section{Sequential Validation and Decision Framework}
\label{sec:sequential}

\subsection{Sequential Reliability Updating}
\label{subsec:sequential_update}

Suppose rollout validation trajectories are generated sequentially under uncertain operating conditions. As additional rollout evidence becomes available, the posterior deployment capability is updated progressively using the accumulated validation outcomes.

Using the Bayesian updating framework developed in Section~\ref{sec:bayesian}, the posterior deployment capability after observing \(n\) rollout validation trajectories is given by \eqref{eq:beta_posterior}. The corresponding posterior approval probability is defined in \eqref{eq:posterior_approval_probability}.

As rollout evidence accumulates, posterior uncertainty is progressively reduced, enabling adaptive approve/reject/continue deployment decisions under finite validation evidence.

\subsection{Sequential Approval and Rejection Rules}
\label{subsec:decision_rule}

The proposed framework introduces a sequential deployment-oriented decision mechanism consisting of three possible actions: approve deployment, reject deployment, and continue validation.

Let \(\tau_A\in(0,1)\) denote the approval confidence threshold and let \(\tau_R\in(0,1)\) denote the rejection confidence threshold. 
At validation stage \(n\), the posterior approval probability is defined in \eqref{eq:posterior_approval_probability}.

The sequential decision rule is defined as follows:
\begin{equation}
	\begin{cases}
		\text{Approve}, 
		& q_n\ge \tau_A,\\
		\text{Reject}, 
		& q_n\le \tau_R\\
		\text{Continue validation}, 
		& \text{otherwise}.
	\end{cases}
	\label{eq:sequential_decision_rule}
\end{equation}

The approval condition indicates that sufficient posterior evidence exists to support deployment reliability above the required threshold \(p_0\). 
Conversely, the rejection condition indicates strong posterior evidence that the controller capability is insufficient.

The intermediate continuation region reflects finite-sample uncertainty. 
When available validation evidence is insufficient to support either approval or rejection with adequate confidence, additional validation trajectories are collected.

The proposed sequential framework differs fundamentally from fixed-sample empirical success-rate evaluation. 
Instead of requiring a predetermined validation size, the framework adaptively determines whether sufficient reliability evidence has been accumulated.

\subsection{Validation Efficiency and Early Stopping}
\label{subsec:efficiency}

One advantage of sequential reliability validation is the possibility of reducing simulation cost through early stopping. 
Highly reliable controllers may achieve approval after relatively few validation trajectories, whereas clearly unreliable controllers may be rejected early without requiring extensive additional simulation.

Let \(N_\pi\) denote the random stopping time of the sequential validation process for policy \(\pi\). 
The stopping time is defined as the first validation stage at which either the approval or rejection condition is satisfied,
\begin{equation}
	N_\pi
	=
	\inf
	\left\{
	n:
	q_n\ge\tau_A
	\text{ or }
	q_n\le \tau_R
	\right\},
	\label{eq:stopping_time}
\end{equation}

Thus, \(N_\pi\) represents the number of rollout validation trajectories required before a deployment decision is reached.

Suppose a conventional fixed-sample validation strategy requires a total of \(N_{\max}\) validation trajectories. 
The expected relative validation saving of the proposed sequential framework is then defined as

\[
\mathrm{Saving}
=
1-
\frac{
	E[N_\pi]
}{
	N_{\max}
}.
\]

The sequential formulation therefore transforms validation into an adaptive reliability-assessment process. 
Controllers with strong reliability evidence may be approved rapidly, whereas uncertain controllers naturally require more extensive validation. The framework therefore allocates rollout-validation effort adaptively according to posterior uncertainty rather than using a fixed validation budget for all controllers \cite{tartakovsky2014sequential,shiryaev2008optimal}.

From a deployment perspective, the framework allocates validation effort adaptively according to posterior uncertainty rather than using identical simulation budgets for all controllers.

\subsection{Deployment-Oriented Interpretation}
\label{subsec:deployment}

The proposed sequential framework may be interpreted as a statistical deployment qualification procedure for learned autonomous controllers. 
The objective is not merely to estimate empirical success frequency, but rather to determine whether the available rollout-validation evidence is sufficient to support deployment under uncertainty. 
Finite validation evidence introduces unavoidable uncertainty regarding the true deployment capability of a learned controller, implying that high empirical success alone does not necessarily guarantee reliable deployment readiness. 
Deployment decisions should therefore depend not only on observed performance, but also on posterior uncertainty regarding unobserved operating conditions and rare failure events.

The proposed framework further separates controller learning from deployment approval. 
The learning stage produces candidate policies, whereas the sequential approval stage evaluates whether the resulting controllers possess sufficiently reliable safety capability under uncertain operating conditions. 
The framework does not constitute formal worst-case verification or certified flight-safety analysis, but instead provides uncertainty-calibrated statistical evidence for deployment-oriented controller evaluation under finite rollout validation conditions. The following sections investigate the proposed framework using autonomous-landing validation experiments under randomized operating conditions and learned control policies.

\section{Simulation-Based Deployment Validation}
\label{sec:simulation}

This section describes the simulation protocol used to evaluate the proposed Bayesian deployment-approval framework. 
The objective of the experiments is not to establish a high-fidelity aircraft model or to optimize reinforcement-learning performance, but to provide a controlled validation setting for studying finite-sample deployment approval under uncertain rollout conditions. 
The predefined reliability and decision parameters used throughout the experiments are summarized in Table~\ref{tab:validation_parameters}.

\begin{table}[!t]
	\centering
	\renewcommand{\arraystretch}{1.0}
	\setlength{\tabcolsep}{4pt}
	\small
	\begin{tabular*}{\columnwidth}{@{\extracolsep{\fill}}ccc}
		\toprule
		Parameter & Value & Description \\
		\midrule
		
		$p_0$
		& 0.95
		& Deployment reliability threshold \\
		
		$\tau_A$
		& 0.95
		& Posterior approval threshold \\
		
		$\tau_R$
		& 0.05
		& Posterior rejection threshold \\
		
		$(\alpha_0,\beta_0)$
		& $(1,1)$
		& Uniform Beta prior hyperparameters \\
		
		$N_{\min}$
		& 30
		& Minimum decision rollout count \\
		
		$N_{\max}$
		& 100
		& Maximum rollout-validation horizon \\
		
		\bottomrule
	\end{tabular*}
	\caption{
		Bayesian deployment-validation parameters used throughout the simulation experiments.
	}
	\label{tab:validation_parameters}
\end{table}

\subsection{Landing Environment and Deployment Conditions}
\label{subsec:simulation_environment}

Simulation experiments are conducted using a simplified two-dimensional autonomous landing environment under randomized operating conditions. 
The environment is intended as a controlled testbed for deployment-oriented reliability assessment rather than as a certified aircraft-dynamics simulator.

The system dynamics, state variables, control formulation, and touchdown safety constraints follow the definitions introduced in Sections~\ref{subsec:environment} and~\ref{subsec:safety}. 
Each rollout episode evolves under randomized initial conditions and stochastic disturbances sampled from the operating-condition distribution discussed in Section~\ref{subsec:uncertainty}. 
The sampled uncertainty includes perturbations in initial altitude, horizontal and vertical velocity, pitch angle, actuator variation, sensor noise, and environmental disturbances.

The landing objective is to guide the vehicle toward the target touchdown region while satisfying the prescribed touchdown safety constraints. 
A rollout episode terminates when safe touchdown is achieved, when instability or crash occurs, or when the episode horizon is exceeded. 
The same deployment-condition distribution is used consistently during validation for all controllers in order to ensure fair posterior reliability comparison.

\subsection{Controller Training and Rollout Validation}
\label{subsec:training_progression}

Learning-based landing controllers are trained under randomized operating conditions using standard reinforcement-learning procedures. 
The main PPO controllers are evaluated across checkpoints from \(1\) million to \(10\) million simulation timesteps, and additional SAC controllers are included to examine whether the observed reward--approval behavior is specific to PPO. 
The reward formulation encourages stable touchdown behavior, trajectory stabilization, attitude regulation, and successful landing completion while penalizing crash events and unstable maneuvers.

Although reward optimization is used to train candidate controllers, deployment validation is performed separately from training. 
For each frozen controller checkpoint, rollout validation trajectories are generated independently under randomized deployment conditions. 
Each validation rollout is converted into the binary safety outcome defined in Section~\ref{subsec:validation}, and the resulting evidence set is used to estimate posterior landing capability and posterior deployment approval confidence using the Bayesian framework developed in Sections~\ref{sec:bayesian} and~\ref{sec:sequential}.

This separation between training and deployment validation is important because training rollouts may provide an optimistic assessment of controller capability. 
Only independent validation rollouts generated after checkpoint freezing are used for posterior deployment approval.

The SAC checkpoints are included as supplementary cross-algorithm validation cases rather than as a full SAC training-progression study. Because the SAC controller reached near-saturated validation performance by approximately \(2\) million training timesteps, additional long-horizon SAC checkpoints were not included in the present experiments.

\subsection{Sequential Deployment Validation Procedure}
\label{subsec:validation_protocol}
\begin{table}[!t]
	\centering
	\renewcommand{\arraystretch}{1.0}
	\setlength{\tabcolsep}{4pt}
	\small
	\begin{tabular*}{\columnwidth}{@{\extracolsep{\fill}}ccc}
		\toprule
		Parameter & Value & Description \\
		\midrule
		$\delta_x$ & 0.20 & Touchdown position tolerance \\
		$\delta_v$ & 0.15 & Vertical touchdown speed tolerance \\
		$\delta_\theta$ & 0.10 & Pitch-angle tolerance \\
		$\delta_{vx}$ & 0.15 & Horizontal touchdown speed tolerance \\
		$\delta_c$ & 0.50 & Minimum touchdown-contact threshold \\
		\bottomrule
	\end{tabular*}
	\caption{
		Touchdown safety thresholds used in rollout validation experiments.
	}
	\label{tab:safety_thresholds}
\end{table}

Deployment validation is performed sequentially for each frozen controller. 
At each validation stage, one additional rollout trajectory is generated under randomized deployment conditions, its binary safety outcome is added to the accumulated evidence set, and the posterior approval probability is updated accordingly. The maximum validation horizon is set to \(N_{\max}=100\) rollout trajectories, which serves as the fixed validation budget in the present simulation study. The touchdown safety thresholds and minimum-evidence safeguard used to compute binary rollout outcomes are summarized in Table~\ref{tab:safety_thresholds}.
For both PPO and SAC checkpoints, the same sequential validation protocol is used such that each frozen controller is evaluated using at most 100 independently generated rollout validation episodes with posterior approval updating performed after every rollout.

The sequential validation framework uses the approve/reject/continue decision mechanism defined in \eqref{eq:sequential_decision_rule}. 
To avoid premature deployment decisions under extremely limited rollout evidence, a minimum-evidence safeguard is additionally imposed such that approval or rejection decisions are not permitted until at least \(N_{\min}=30\) rollout validation trajectories have been observed. Sequential validation terminates once the approval condition, rejection condition, or maximum validation horizon is reached.

Controllers with strong posterior evidence of insufficient deployment capability may therefore be rejected early, whereas near-boundary controllers continue accumulating additional rollout evidence until sufficient posterior confidence is obtained or the validation horizon is exhausted. 
The following experiments investigate sequential posterior evolution, reward--reliability mismatch, finite-sample approval conservatism, and deployment-confidence progression across PPO and SAC controller checkpoints. The values of \(N_{\min}=30\) and \(N_{\max}=100\) are experimental protocol settings used to illustrate finite-sample deployment approval behavior under a controlled validation budget and are not intended as universal certification requirements. Larger validation horizons would generally produce stronger posterior concentration and potentially different approval outcomes near the deployment boundary. 

\section{Experimental Results}
\label{sec:results}

\subsection{Sequential Bayesian Approval Dynamics}
\label{subsec:posterior_evolution}

The first experiment investigates the sequential evolution of posterior deployment approval probability under progressively increasing rollout validation evidence. 
For each learned controller, validation trajectories are generated sequentially under randomized deployment conditions, and the posterior approval probability in \eqref{eq:posterior_approval_probability} is updated continuously as additional rollout evidence becomes available.

\begin{figure*}[!t]
	\centering
	
	\subfloat[Sequential Bayesian approval evolution during rollout validation.]{
		\label{fig:posterior_evolution}
		\includegraphics[width=0.46\textwidth]{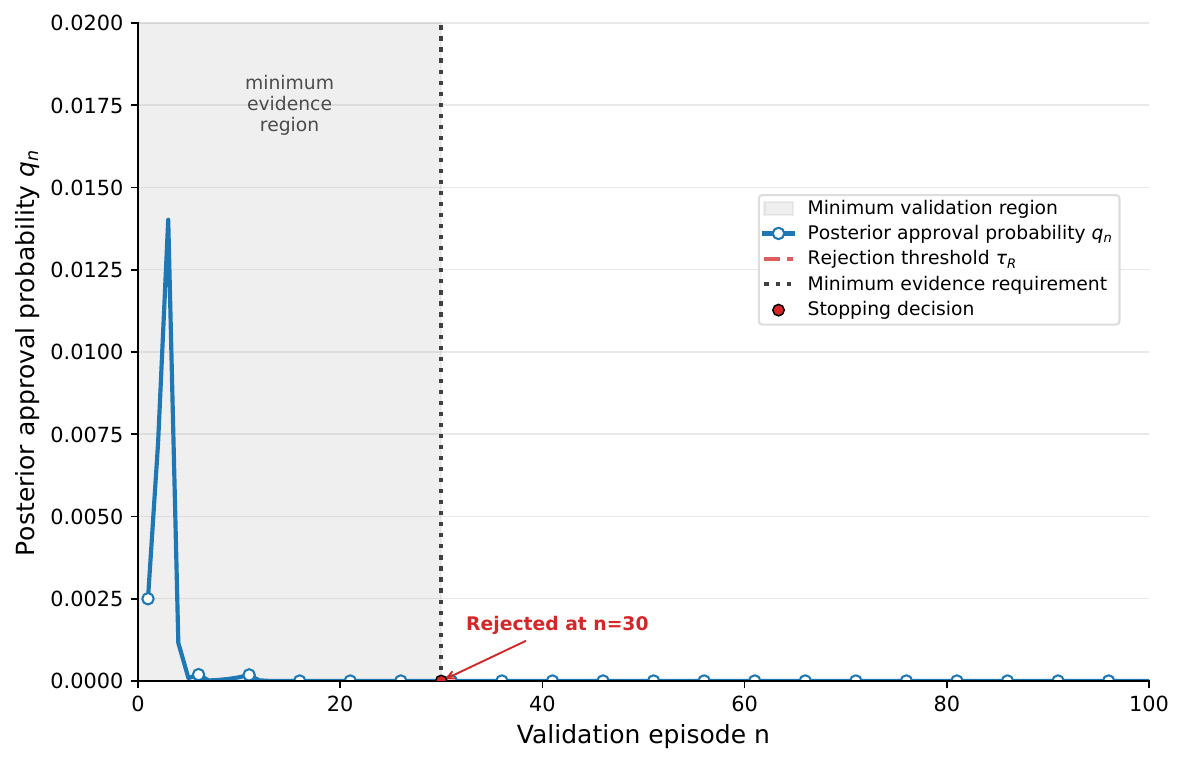}
	}
	\hspace{4pt}
	\subfloat[Reward--safety distributions across PPO checkpoints.]{
		\label{fig:reward_reliability}
		\includegraphics[width=0.46\textwidth]{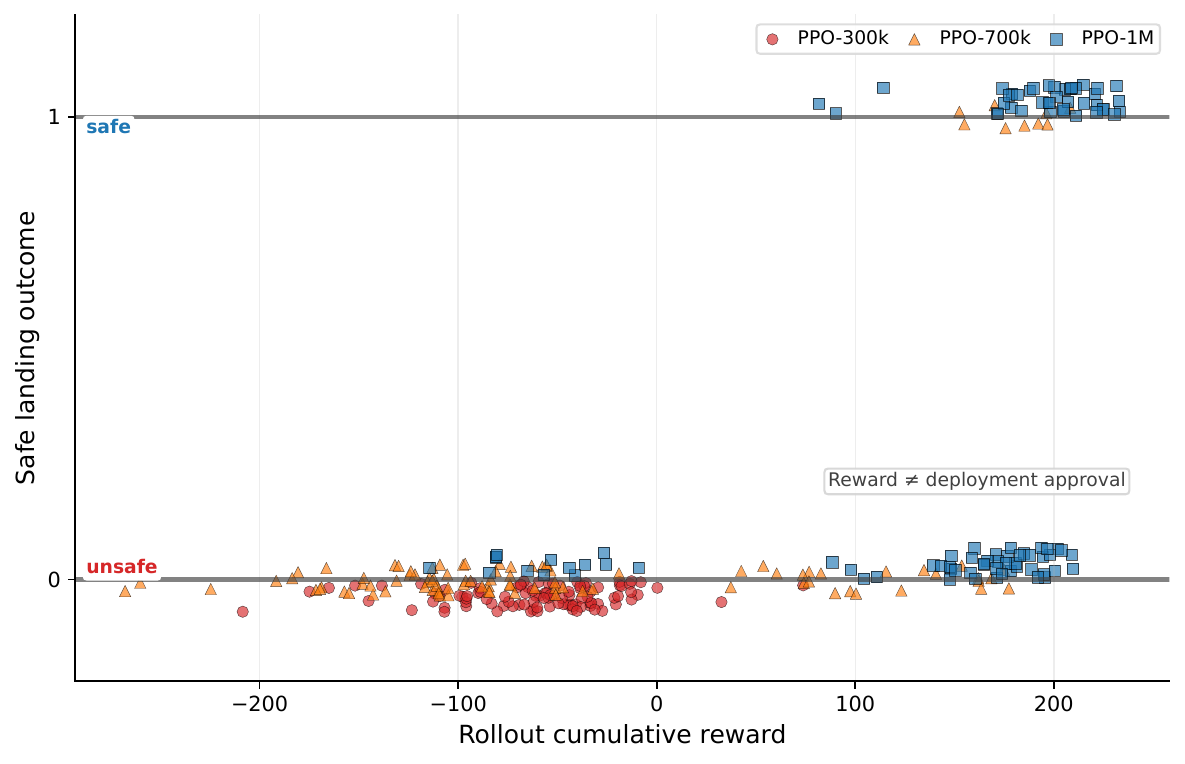}
	}
	\caption{
		Sequential deployment-validation behavior and reward--approval mismatch for learned landing controllers. 
		Panel (a) illustrates sequential Bayesian approval evolution during rollout validation. 
		The posterior approval probability is updated continuously as rollout evidence accumulates, while the minimum-evidence safeguard prevents premature deployment decisions. 
		The small vertical scale reflects rapid decrease toward a near-zero posterior approval regime under the specified deployment reliability requirement, resulting in early rejection after the minimum-evidence stage.
		Panel (b) compares rollout cumulative reward and safe landing outcomes across PPO checkpoints. 
		Although cumulative reward generally improves during training, posterior deployment approval evolves substantially more conservatively, illustrating that reward optimization and deployment readiness represent different evaluation objectives.
	}
	\label{fig:deployment_validation_results}
\end{figure*}

The first row of Fig.~\ref{fig:deployment_validation_results} summarizes the resulting posterior evolution behavior together with the reward--reliability relationship across PPO checkpoints.

Several important observations emerge from Fig.~\ref{fig:deployment_validation_results}(a). 
First, posterior approval confidence evolves dynamically rather than remaining fixed throughout validation. 
Second, finite validation evidence produces substantial uncertainty during early rollout stages, motivating the minimum-evidence safeguard introduced in Section~\ref{subsec:decision_rule}. 
Third, deployment rejection may occur substantially earlier than fixed-horizon validation once sufficient posterior evidence against deployment reliability has accumulated. The posterior approval probability decreases rapidly after several unsafe rollout observations, leading to early rejection once sufficient evidence against deployment reliability accumulates.

The results therefore demonstrate that deployment validation should be interpreted as a sequential statistical inference process rather than a static empirical success-rate evaluation problem. 
Even when early rollout trajectories appear favorable, substantial uncertainty regarding the true deployment capability may remain under finite validation evidence.

\begin{table}[!t]
	\centering
	\renewcommand{\arraystretch}{1.0}
	\setlength{\tabcolsep}{4pt}
	\small
	\begin{tabular*}{\columnwidth}{@{\extracolsep{\fill}}ccc}
		\toprule
		Controller & Stopping rollout & Final decision \\
		\midrule
		
		PPO-1M  & 30  & Reject \\
		PPO-2M  & 32  & Reject \\
		PPO-3M  & 30  & Reject \\
		PPO-4M  & 30  & Reject \\
		PPO-5M  & 30  & Reject \\
		PPO-6M  & 31  & Reject \\
		PPO-7M  & 100 & Continue \\
		PPO-8M  & 30  & Reject \\
		PPO-9M  & 31  & Reject \\
		PPO-10M & 100 & Continue \\
		
		\midrule
		
		SAC-1M  & 30  & Reject \\
		SAC-2M  & 74  & Approve \\
		
		\bottomrule
	\end{tabular*}
	\caption{
		Sequential deployment-validation outcomes across PPO and SAC controller checkpoints under the proposed Bayesian approval framework. 
		Controllers with clearly insufficient deployment capability are rejected relatively early once sufficient posterior evidence against deployment approval accumulates, whereas near-boundary controllers naturally require substantially larger rollout-validation budgets before reliable decisions can be reached.
	}
	\label{tab:sequential_validation_outcomes}
\end{table}
Table~\ref{tab:sequential_validation_outcomes} provides an operational illustration of the sequential deployment-validation behavior across PPO and SAC controller checkpoints. Controllers with clearly insufficient deployment capability are rejected relatively early once sufficient posterior evidence against deployment approval accumulates after the minimum-evidence safeguard stage. In contrast, near-boundary controllers such as PPO-7M and PPO-10M remain within the continuation region throughout the rollout horizon despite relatively strong empirical landing success, reflecting persistent posterior uncertainty regarding whether the true deployment capability satisfies the required reliability target. The stronger SAC-2M controller eventually satisfies the deployment approval condition before reaching the maximum rollout horizon, illustrating adaptive rollout allocation under the proposed Bayesian framework.

\subsection{Reward--Reliability Mismatch}
\label{subsec:reward_reliability_text}

Figure~\ref{fig:deployment_validation_results}(b) investigates the relationship between reinforcement learning reward optimization and deployment-oriented landing reliability.

The figure demonstrates that cumulative reward alone does not provide sufficient evidence for deployment readiness. 
Several controllers achieve relatively large cumulative reward values while still producing unsafe touchdown trajectories under uncertain deployment conditions. 
Unsafe landing outcomes remain observable even among rollouts associated with comparatively strong reward performance.

These observations highlight an important limitation of conventional reinforcement learning evaluation. 
Expected reward optimization primarily measures average control performance, whereas approval requires statistical confidence regarding safety behavior under uncertainty. 
Consequently, reward progression and reliability should not be interpreted as equivalent quantities.

The results therefore support the central motivation of the proposed Bayesian framework: deployment-oriented controller validation requires uncertainty-calibrated reliability assessment rather than reward evaluation alone.

\subsection{Finite-Sample Approval Conservatism}
\label{subsec:approval_boundary}

The third experiment investigates the effect of finite validation evidence on approval conservatism.

\begin{figure*}[!t]
	\centering
	
	\subfloat[Bayesian approval boundary under finite validation evidence.]{
		\label{fig:approval_boundary}
		\includegraphics[width=0.46\textwidth]{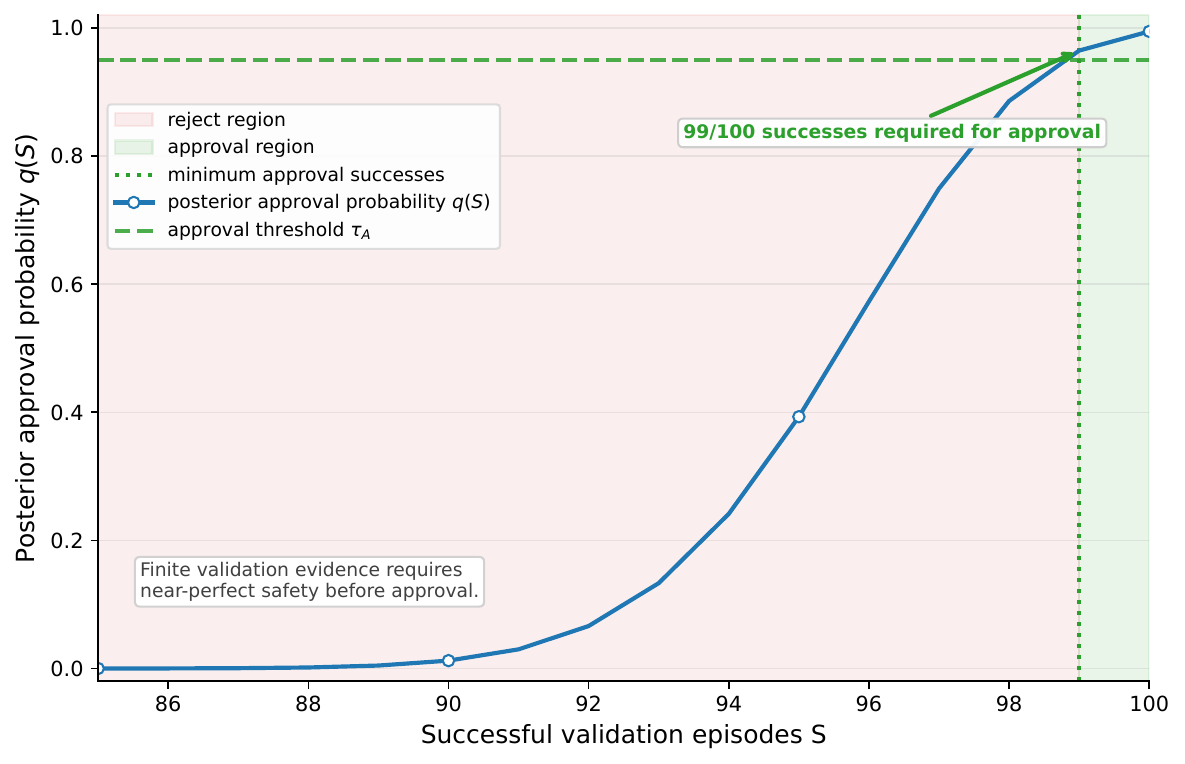}
	}
	\hspace{4pt}
	\subfloat[Training progression toward deployment approval.]{
		\label{fig:training_progression}
		\includegraphics[width=0.46\textwidth]{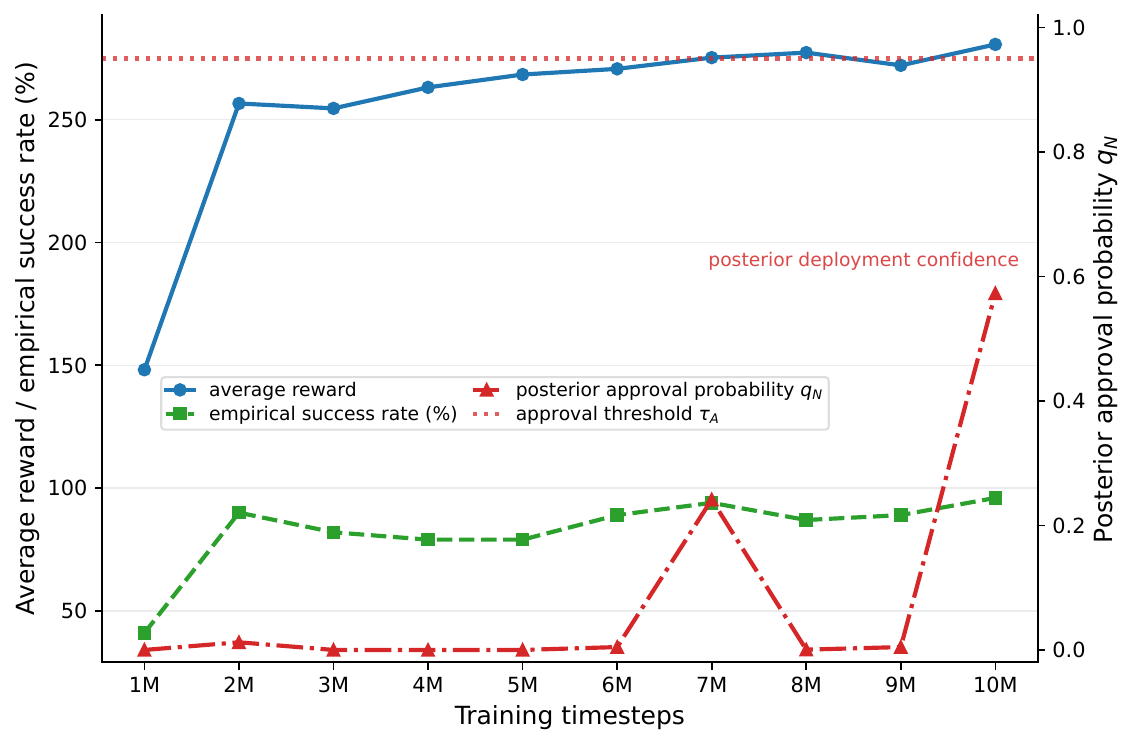}
	}
	\caption{
		Finite-sample approval conservatism and deployment-confidence progression during PPO training. 
		Panel (a) illustrates the Bayesian approval boundary under finite rollout validation evidence. 
		Near-perfect empirical landing success may still be insufficient for deployment approval due to posterior uncertainty regarding the true deployment capability. 
		Panel (b) summarizes reward progression, empirical landing success progression, and posterior approval-confidence evolution across PPO training stages from \(1\) million to \(10\) million simulation timesteps. 
		Although reward and empirical success improve rapidly during training, posterior approval confidence evolves substantially more conservatively and exhibits noticeable fluctuations near the approval boundary, demonstrating that reward optimization and deployment approval progress at substantially different rates.
	}
	\label{fig:approval_progression_results}
\end{figure*}

For a maximum rollout-validation size of \(N_{\max}=100\), the posterior approval probability is evaluated as a function of the number of successful rollout validation episodes. The second row of Fig.~\ref{fig:approval_progression_results} summarizes the resulting finite-sample approval behavior together with deployment-confidence progression across PPO training stages.

Figure~\ref{fig:approval_progression_results}(a) demonstrates that finite-sample uncertainty induces highly conservative approval behavior. 
Near-perfect empirical landing success may still be insufficient for deployment approval because posterior approval confidence remains sensitive to finite rollout uncertainty regarding the true deployment capability.

This result highlights a fundamental distinction between empirical landing success frequency and posterior approval confidence. For example, observing \(90\%\) empirical landing success may still correspond to substantial uncertainty regarding whether the true deployment capability satisfies the required safety target in \eqref{eq:deployment_requirement}.

Consequently, controllers exhibiting apparently strong empirical performance may nevertheless remain statistically uncertain with respect to deployment readiness.

The figure therefore illustrates that approval is fundamentally a statistical confidence problem under finite validation evidence rather than a direct thresholding operation on empirical success frequency.

\begin{table}[!t]
	\centering
	\renewcommand{\arraystretch}{1.0}
	\setlength{\tabcolsep}{3.2pt}
	\small
	\begin{tabular*}{\columnwidth}{@{\extracolsep{\fill}}lcccc}
		\toprule
		Controller 
		& Success 
		& Empirical 
		& $q_N$ 
		& Bayesian \\
		\midrule
		
		PPO-1M  & 41.3\% & Reject  & 0.0000 & Reject \\
		PPO-2M  & 89.8\% & Reject  & 0.0123 & Reject \\
		PPO-3M  & 82.1\% & Reject  & 0.0000 & Reject \\
		PPO-4M  & 78.9\% & Reject  & 0.0000 & Reject \\
		PPO-5M  & 79.4\% & Reject  & 0.0000 & Reject \\
		PPO-6M  & 88.6\% & Reject  & 0.0046 & Reject \\
		PPO-7M  & 94.1\% & Reject  & 0.2415 & Continue \\
		PPO-8M  & 87.2\% & Reject  & 0.0005 & Reject \\
		PPO-9M  & 88.9\% & Reject  & 0.0046 & Reject \\
		PPO-10M & 95.7\% & Approve & 0.5729 & Continue \\
		
		\midrule
		
		SAC-1M  & 88.0\% & Reject  & 0.0016 & Reject \\
		SAC-2M  & 99.0\% & Approve & 0.9645 & Approve \\
		
		\bottomrule
	\end{tabular*}
	\caption{
		Summary of empirical landing success and posterior deployment approval confidence across PPO and SAC controller checkpoints under the proposed Bayesian deployment-validation framework. 
		The empirical approval rule approves deployment whenever the observed landing success rate exceeds the required deployment threshold \(p_0=0.95\), without accounting for finite-sample uncertainty. 
		In contrast, the Bayesian framework incorporates posterior uncertainty regarding the true deployment capability under finite rollout validation evidence. 
		The results illustrate that empirical success alone may lead to overconfident deployment interpretation, whereas posterior approval inference remains substantially more conservative near the deployment boundary.
	}
	\label{tab:training_progression_summary}
\end{table}

Table~\ref{tab:training_progression_summary}
summarizes empirical landing success and posterior deployment approval confidence across PPO and SAC controller checkpoints under both conventional empirical-success approval and the proposed Bayesian deployment-validation framework.

The results reveal a substantial discrepancy between empirical-success-based deployment interpretation and uncertainty-calibrated Bayesian approval. 
Under the conventional empirical rule, deployment is approved whenever the observed landing success rate exceeds the required threshold \(p_0=0.95\), corresponding to a direct plug-in thresholding decision based solely on the empirical estimator \(\hat p_N\). 
However, the Bayesian framework additionally accounts for finite-sample uncertainty regarding the unknown deployment capability.

Thus, even when the empirical estimate slightly exceeds the required capability threshold, the posterior probability that the true deployment capability exceeds the threshold may remain substantially below the required approval confidence level.

Several PPO checkpoints achieve relatively strong empirical landing success while still remaining statistically uncertain under Bayesian approval. For example, PPO-10M achieves approximately \(95.7\%\) empirical landing success and would therefore be approved under the conventional empirical rule. Nevertheless, the posterior approval probability remains only \(q_N=0.5729\), indicating substantial residual uncertainty regarding whether the true deployment capability satisfies the required reliability target under finite rollout validation evidence. The observed conservatism is a direct consequence of finite-sample uncertainty near the deployment boundary under stringent reliability requirements. Similarly, PPO-7M achieves approximately \(94\%\) empirical landing success while remaining far below the Bayesian approval threshold.

The supplementary SAC experiments further demonstrate that the discrepancy between empirical success and Bayesian approval is not specific to PPO controllers. 
For example, SAC-1M achieves approximately \(88\%\) empirical landing success while remaining strongly rejected under Bayesian approval due to insufficient posterior deployment confidence. 
After further training, SAC-2M enters a deployable regime in which both empirical landing success and posterior approval confidence exceed the required deployment threshold.

Overall, the results demonstrate that empirical success frequency alone may produce overconfident deployment interpretation under finite rollout validation evidence. 
The proposed Bayesian framework instead provides a more conservative and uncertainty-calibrated deployment assessment by explicitly incorporating posterior uncertainty regarding the true deployment capability.

\subsection{Training Progression Toward Deployment Approval}
\label{subsec:training_progression_text}

Figure~\ref{fig:approval_progression_results}(b) investigates the evolution of reward performance, empirical landing success, and posterior approval confidence across PPO training progression.

PPO controllers are frozen periodically throughout training from \(1\) million to \(10\) million simulation timesteps and evaluated independently under rollout validation conditions.

The figure shows that cumulative reward and empirical landing success improve substantially during PPO training. 
However, posterior approval confidence evolves considerably more conservatively under finite validation evidence. 
Controllers achieving approximately \(90\%\) empirical landing success may still exhibit extremely small posterior approval probability due to insufficient statistical confidence regarding the underlying reliability.

Another important observation is the presence of non-monotonic posterior fluctuations near the approval boundary. 
Although empirical landing success remains relatively stable for stronger controllers, posterior confidence may fluctuate noticeably under finite rollout validation. 
This behavior reflects the sensitivity of Bayesian approval inference when the controller operates near the required reliability threshold.

Overall, the results demonstrate that reward optimization, empirical success, and posterior approval confidence evolve at substantially different rates during controller training. 
The proposed framework therefore provides a statistically grounded mechanism for uncertainty-aware deployment validation under finite observations.

\section{Discussion}
\label{sec:discussion}

\subsection{Deployment Approval under Finite Validation Uncertainty}
\label{subsec:discussion_performance}

The proposed framework highlights an important distinction between empirical controller performance and deployment-oriented approval inference. 
Conventional reinforcement learning evaluation primarily emphasizes cumulative reward and empirical success frequency, yet these quantities alone do not fully characterize uncertainty regarding the true deployment capability of learned controllers under finite rollout validation evidence. 
The present work instead reformulates learned-controller evaluation as a probabilistic deployment-approval problem, where posterior inference is used to quantify uncertainty regarding whether the underlying landing capability satisfies the required deployment target under uncertain operating conditions.

The results demonstrate that reward optimization and deployment approval are fundamentally different objectives. 
Policies achieving high cumulative reward or strong empirical landing success may still exhibit insufficient deployment confidence when validation evidence remains limited or when rare unsafe events are insufficiently sampled. 
The resulting posterior approval probabilities therefore provide uncertainty-calibrated deployment assessment rather than deterministic certification. 
The sequential validation framework further highlights the adaptive nature of deployment approval, where controllers exhibiting strong statistical evidence of reliable behavior may be approved rapidly, whereas uncertain controllers naturally require additional rollout validation before reliable deployment decisions can be made. 
More broadly, the proposed framework provides a statistical connection between conventional reinforcement-learning evaluation and deployment-oriented validation under finite-sample uncertainty.

\subsection{Operating-Distribution Dependence}
\label{subsec:discussion_distribution}

The proposed landing capability formulation is inherently dependent on the operating-condition distribution used during validation. 
Controller performance is evaluated relative to the disturbance distribution, environmental uncertainty, and initial-condition variability assumed during rollout generation.

Consequently, posterior approval confidence should not be interpreted as a universal property independent of the deployment environment. 
A controller validated under one operating-condition distribution may exhibit substantially different capability under shifted environmental conditions or previously unseen disturbances.

This observation is particularly important for rollout-based autonomous-system validation.
Simulation environments inevitably simplify real-world operating conditions, and validation distributions may not perfectly represent deployment environments. 
The proposed framework therefore evaluates statistical deployment confidence conditioned on the assumed validation distribution rather than universal worst-case safety.

Nevertheless, the framework remains valuable because it provides explicit probabilistic interpretation of deployment capability under clearly specified uncertainty assumptions. 
The operating-condition distribution can also be progressively expanded to include increasingly challenging disturbances, thereby enabling robustness-oriented validation analysis.

\subsection{Limitations and Future Extensions}
\label{subsec:discussion_limitations}

Several limitations of the present work should be acknowledged.

First, the simulation environment adopts a simplified low-fidelity landing model intended primarily for methodological investigation of finite-sample approval inference. 
The present work does not attempt to reproduce certified aircraft dynamics, high-fidelity aerodynamics, or flight-certified control systems. 
Instead, the environment serves as a controlled platform for studying statistical deployment uncertainty under uncertain operating conditions.

Second, the proposed framework does not attempt to provide formal worst-case verification or deterministic certification guarantees. 
Instead, the framework focuses on probabilistic approval inference under finite validation evidence and specified operating-condition distributions.

Third, the Bernoulli validation model assumes conditionally independent rollout validation trajectories under the specified operating-condition distribution. In practical autonomous systems, correlated disturbances, temporal dependencies, or nonstationary environments may introduce additional complexity into posterior capability estimation.

The present experiments primarily investigate deployment-oriented posterior behavior and sequential decision dynamics rather than formal posterior calibration under repeated Monte Carlo replication. Large-scale calibration analysis of posterior approval probabilities and empirical false-approval frequencies remains an important direction for future investigation.

Several important future extensions remain possible. 
The proposed framework may be extended toward distribution-shift robustness analysis, sim-to-real uncertainty quantification, rare-event reliability estimation, adaptive validation-budget allocation, and reliability-aware policy selection across multiple learned controllers \cite{zhao2020sim}. 

The proposed framework may also complement runtime safety-monitoring and online intervention mechanisms by providing uncertainty-calibrated pre-deployment statistical approval before operational rollout. The framework may also be generalized beyond autonomous landing to broader classes of learned autonomous systems, including robotic manipulation, autonomous driving, and safety-critical embodied AI systems.

Overall, the present work demonstrates that finite-sample probabilistic approval inference provides a useful complementary perspective to conventional reward-based evaluation of learned autonomous controllers.
\section{Conclusion}
\label{sec:conclusion}

This work developed a Bayesian deployment approval framework for learned autonomous landing controllers under finite rollout validation evidence. Rather than evaluating controllers solely through cumulative reward or empirical success frequency, the proposed framework reformulates deployment-oriented controller validation as a probabilistic approval inference problem under uncertainty. A probabilistic landing capability formulation was introduced together with Bayesian posterior approval inference and sequential approve/reject/continue validation under finite rollout evidence.

Simulation experiments demonstrated that empirical success frequency and cumulative reward may produce misleading deployment interpretation under limited validation samples. The results further showed that reward optimization, empirical landing success, and posterior approval confidence may evolve at substantially different rates during reinforcement learning training, while controllers exhibiting strong empirical performance may nevertheless remain statistically uncertain with respect to deployment readiness near the approval boundary. The proposed posterior approval framework therefore provides a more uncertainty-calibrated interpretation of deployment readiness under finite validation evidence.

The framework is intentionally independent of the underlying controller architecture and may therefore be applied to reinforcement learning policies, heuristic controllers, classical feedback systems, or hybrid autonomous-control frameworks. Although the present study focused on autonomous landing environments, the formulation may be generalized to broader classes of learned autonomous systems involving uncertainty-sensitive deployment decisions. Future work may extend the framework toward hierarchical Bayesian approval formulations, distribution-shift robustness analysis, rare-event reliability estimation, and sim-to-real uncertainty quantification for safety-critical autonomous systems operating under uncertainty. More broadly, the present work suggests that deployment readiness for learned autonomous systems should be interpreted not only through empirical performance, but also through uncertainty-aware statistical evidence under finite validation conditions.


\section*{DECLARATION}
\begin{description}
	\item[Funding:] Not applicable. This research did not receive any specific grant from funding agencies in the public, commercial, or not-for-profit sectors.
	\item[Conflicts of interest / Competing interests:] The authors declare that they have no conflicts of interest or competing interests.
	\item[Availability of data and material:] Not applicable.
	\item[Code availability:] Not applicable.
	\item[Ethics approval:] Not applicable.
	\item[Consent to participate:] Not applicable.
	\item[Consent for publication:] All authors consent to the publication of this work and approve the final version of the manuscript.
\end{description}

\bibliographystyle{unsrtnat}
\bibliography{references}

\end{document}